\pdfoutput=1

\documentclass[11pt]{article}

\usepackage{EMNLP2022}
\usepackage{bm}
\usepackage{times}
\usepackage{latexsym}
\usepackage{graphicx}
\usepackage{multicol}
\usepackage{multirow}
\usepackage{amsmath}
\usepackage{arydshln}
\usepackage{algorithm}
\usepackage{algpseudocode}
\usepackage{pgfplots}
\usepackage{enumerate}
\usepackage{enumitem}
\usepackage{amsfonts,amssymb}
\usepackage[T1]{fontenc}

\usepackage[utf8]{inputenc}

\usepackage{microtype}

\usepackage{inconsolata}

\usepackage{CJKutf8}

\usepackage{enumitem}
\setenumerate[1]{itemsep=0pt,partopsep=0pt,parsep=\parskip,topsep=5pt}
\setitemize[1]{itemsep=0pt,partopsep=0pt,parsep=\parskip,topsep=5pt}
\setdescription{itemsep=0pt,partopsep=0pt,parsep=\parskip,topsep=5pt}

\title{Entity-centered Cross-document Relation Extraction}

\author{
Fengqi Wang\textsuperscript{\rm 1}, \,
Fei Li\textsuperscript{\rm 1}\Thanks{\ \ Corresponding author} ,  \,
Hao Fei\textsuperscript{\rm 2},  \,
Jingye Li\textsuperscript{\rm 1},  \,
Shengqiong Wu\textsuperscript{\rm 2}, \\
\textbf{
Fangfang Su\textsuperscript{\rm 1},  \,
Wenxuan Shi\textsuperscript{\rm 1},  \,
Donghong Ji\textsuperscript{\rm 1} and Bo Cai\textsuperscript{1*}
}\\
\textsuperscript{\rm 1} Key Laboratory of Aerospace Information Security and Trusted Computing, Ministry  \\of Education,
School of Cyber Science and Engineering, Wuhan University, China \\
\textsuperscript{\rm 2} School of Computing, National University of Singapore, Singapore\\
\texttt{\{wangfengqi,lifei\_csnlp,theodorelee,fangsu,shiwenxuan,dhji,caib\}@whu.edu.cn} \\
\texttt{haofei37@nus.edu.sg, swu@u.nus.edu}
}

\begin{document}
\begin{CJK}{UTF8}{gbsn}
\maketitle
\begin{abstract}
Relation Extraction (RE) is a fundamental task of information extraction, which has attracted a large amount of research attention. Previous studies focus on extracting the relations within a sentence or document, while currently researchers begin to explore cross-document RE.
However, current cross-document RE methods directly utilize text snippets surrounding the target entities in multiple given documents, which brings considerable noisy and non-relevant sentences.
Moreover, they utilize all the text paths in a document bag in a coarse-grained way, without considering the connections between these text paths.
In this paper, we aim to address both of these shortages and push the state-of-the-art for cross-document RE.
First, we focus on input construction for our RE model and propose an entity-based document-context filter to retain useful information in the given documents by using the bridge entities in the text paths.
Second, we propose a cross-document RE model based on cross-path entity relation attention, which allows the entity relations across text paths to interact with each other.
We compare our cross-document RE method with the state-of-the-art methods in the dataset CodRED.
Our method outperforms them by at least  10\% in F1, thus demonstrating its effectiveness.\footnote{Code: \href{https://github.com/MakiseKuurisu/ecrim}{https://github.com/MakiseKuurisu/ecrim}}

\end{abstract}

\section{Introduction}

Relation Extraction (RE) aims to detect the semantic relations between a pair of target entities in a given text, which has long been a fundamental task in natural language processing (NLP).
Most of RE studies are under the assumption that entity pairs are within a sentence (i.e., sentence-level RE) \cite{zeng-etal-2014-relation, dos-santos-etal-2015-classifying,cai-etal-2016-bidirectional,zhou-etal-2016-attention,zhang-etal-2018-graph,FeiZRJ21} or a document (i.e., document-level RE) \cite{christopoulou-etal-2019-connecting, nan-etal-2020-reasoning, zeng-etal-2020-double,li-etal-2021-mrn,FeiDiaREIJCAI22,DBLP:conf/ijcai/ZhangCXDTCHSC21}.
Another line considers the research of cross-text RE, where entity pairs are separated into different text units, (i.e., cross-sentence RE or N-ary RE) \cite{PengPQTY17}.

\begin{figure*}[ht]
    \centering
    \includegraphics[width=1\textwidth,height=0.32\textwidth]{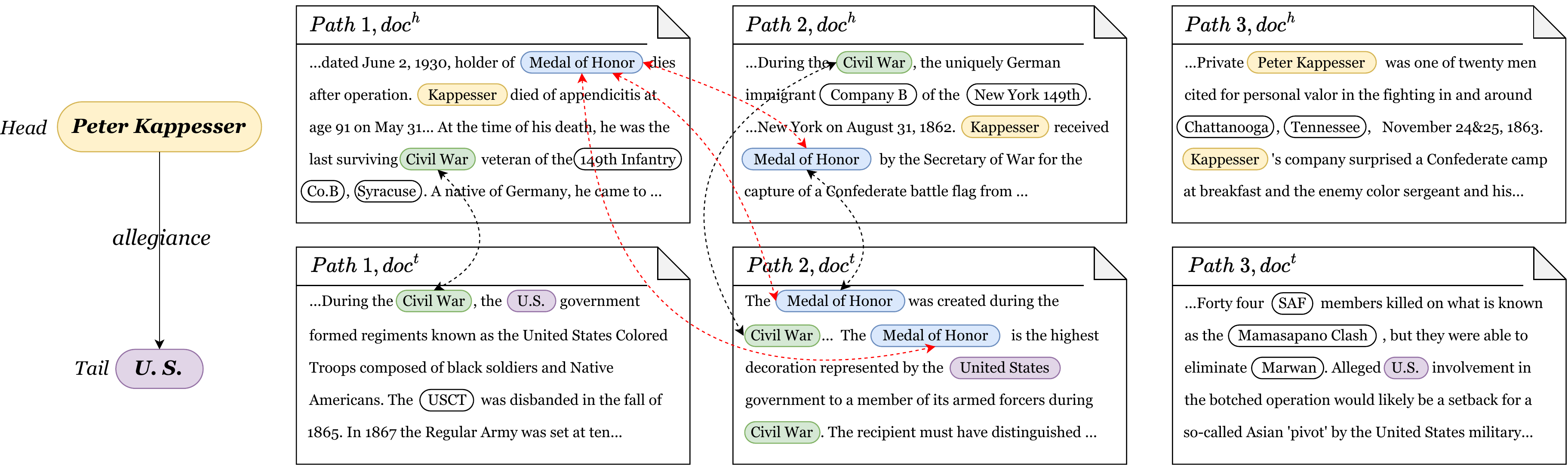}
    \caption{
    An example to show the setting of cross-document RE.
    In this document bag, there are three text paths to imply the \emph{allegiance} relation between the head entity \emph{Peter Kappesser} and tail entity \emph{U.S.}.
    Each text path has two documents, where one contains the head entity and the other one contains the tail entity.
    In each text path, the head and tail entities are bridged by another entity appearing in both documents (e.g., \emph{Civil War}).
    }
    \label{fig:example}
    \vspace{-9pt}
\end{figure*}

The latest RE research has moved to cross-document RE (CodRE), i.e., the target entities are located in different documents \cite{yao-etal-2021-codred}.
As exemplified in Fig. \ref{fig:example}, a CodRE model needs to first retrieve the relevant documents and then recognizes the key \textbf{text paths} in these documents for relation reasoning.
In \citet{yao-etal-2021-codred}, the task is formalized based on the idea of distant supervision \cite{mintz-etal-2009-distant}, i.e., the text paths in a bag can facilitate the relation reasoning and thus their model performs bag-level prediction over all the text paths.
Unfortunately, their method may suffer from at least two problems, which inevitably hinder the accurate relation inference.

First, the inputs of their method are not tailor-made for cross-document RE.
For instance, they extract text snippets surrounding the two target entities in the document as input of a bag, brings much noisy and non-relevant context information. 
Moreover, they ignore important bridge entities in the text paths of the bag, leading to the loss of instructive and salience information for cross-document RE. 
As can be seen in Fig. \ref{fig:example}, the sentences containing bridge entities are necessary to reason the relations between target entities and missing them will seriously affect the reasoning process.

Second, their method does not make full use of the connections between text paths.
For example, the pipeline model proposed by \citet{yao-etal-2021-codred} simply leverages the information of the text path in an isolated way, lacking deep consideration of the global connections of all text paths. 
In contrast, although their end-to-end model \cite{yao-etal-2021-codred} uses the context of all the text paths, the process of synthesizing the context is coarse-grained.
The connections across multiple text paths are actually beneficial for cross-document RE.
As shown in Fig. \ref{fig:example}, the entity ``\emph{Medal of Hornor}'' provides an additional link for different text paths, which helps to reason the ``\emph{allegiance}'' relation between ``\emph{Peter Kappesser}'' and ``\emph{U.S.}''.

Therefore, in this paper, we focus on addressing the above problems and improving the performance of cross-document RE by presenting a novel \textbf{E}ntity-based \textbf{C}ross-path \textbf{R}elation \textbf{I}nference \textbf{M}ethod (\textbf{ECRIM}).
First, we propose an \textbf{entity-based document-context filter} to elaborately construct the input for our cross-document RE model, which includes two steps:
\textbf{1)} We filter out a number of sentences based on their scores with regards to bridge entities. Three heuristic conditions are used to describe the importance scores of bridge entities and then these scores are assigned to the sentences for filtering.
\textbf{2)} After filtering out the sentences with lower scores, we use the semantic-based sentence filter to reorder the remaining sentences, making them into a relatively coherent document, inspired by the method of sentence ordering in multi-document summarization. \cite{DBLP:journals/corr/abs-1106-1820,DBLP:journals/corr/abs-1909-10393}.

After input construction, we propose a novel cross-document RE model that is equipped with a \textbf{cross-path entity relation attention} module to capture the connections of text paths within a document bag, inspired by \citet{10.1145/1553374.1553534,DBLP:journals/corr/abs-1906-04881}.
Specifically, we build a relation matrix where each unit represents a relation between two entities belonging to the same bag. 
Then the bag-level relation matrix is able to capture the dependencies between the relations by the attention mechanism \cite{NIPS2017_3f5ee243}, which allows one relation to focus on other more relevant relations in the text paths by modeling the discourse structure \cite{FeiMatchStruICML22,0001RJ20a,Wu0LZLTJ22}.

We conduct experiments on the CodRED dataset \cite{yao-etal-2021-codred}. The results show that our model outperforms the baseline models by a large margin. In summary, our contributions can be summarized as follows:

\begin{itemize}
\item We apply an entity-based document-context filter to retain useful context information and important bridge entities across the documents.

\item We propose a cross-path entity relation attention model for cross-document RE, which allows the relation representations across text paths to interact with each other respect to bridge entities.


\item We validate the effectiveness of our model, which significantly pushes the state-of-the-art performance for cross-document RE.

\end{itemize}

\begin{figure*}[!t]
    \centering
    \includegraphics[width=1\textwidth]{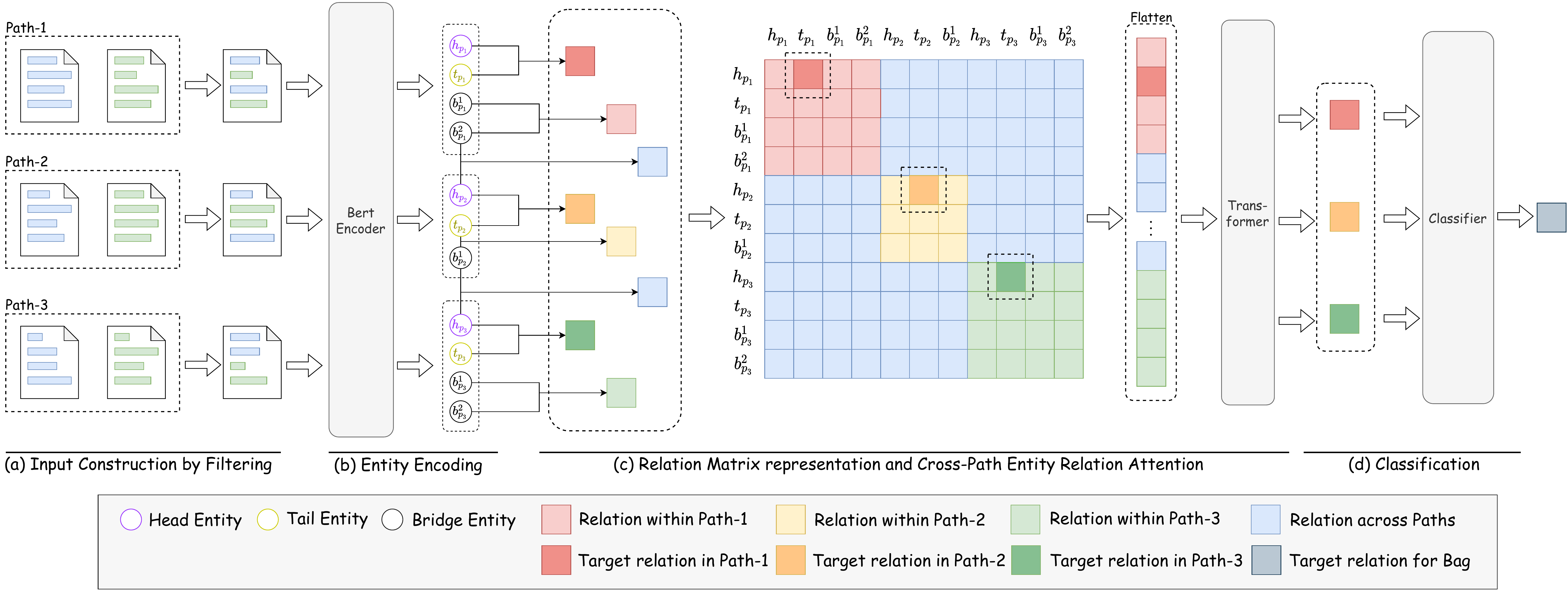}
    \caption{
    The overall architecture of our system.
    (a) utilizes a entity-based document-context filter to select the sentences that are relevant to the target entity pair (cf. Section \ref{sec:entity_document_filter}). (b) yields entity embeddings from contextualized word representations (cf. Section \ref{sec:encoder}). (c) leverages the cross-path entity relation attention to capture the connections between the entities and relations of all the paths in the bag (cf. Section \ref{sec:cross_path_attention}). 
    (d) aggregates the predictions of all the paths to get a bag-level prediction.
    }
    \label{fig:framework}
    \vspace{-8pt}
\end{figure*}

\section{Related Work}

\subsection{Sentence-level Relation Extraction}
Relation Extraction is one of the key tasks of information extraction community \cite{RenFP18,FeiCrossACL20,FeiGraphSynAAAI21,0001JLLRL21,cao-etal-2022-oneee,Li00WZTJL22}.
Sentence-level RE aims at identifying the relationship between two entities in a sentence and many efforts have been devoted to this problem.
\citet{zeng-etal-2014-relation} exploit a convolutional deep neural network to extract lexical and sentence-level features.
\citet{dos-santos-etal-2015-classifying} tackle the sentence-level RE task by using a convolutional neural network that performs classification by ranking.
\citet{cai-etal-2016-bidirectional} present a novel model that strengthen the ability to classifying directions of relationships between entities.
\citet{zhang-etal-2018-graph} propose an extension of graph convolutional networks \cite{MMGCN19,GRCN20} and applied a novel pruning strategy to incorporate relevant information while removing irrelevant content.

\subsection{Document-level Relation Extraction}
Recent years, researchers have shown a growing interest for document-level text mining \cite{0001RWLJ21,FeiWRLJ21,zhang-etal-2021-multi,yang-etal-2021-document}.
Document-level RE aims to detect the relations within one document.
\citet{christopoulou-etal-2019-connecting} utilize different types of nodes and edges to create a document-level graph for document-level RE.
\citet{nan-etal-2020-reasoning} propose a novel model that empowers the relational reasoning across sentences by automatically inducing the latent document-level graph.
\citet{zeng-etal-2020-double} propose a graph aggregation and inference network to infer relations between entities across long paragraphs.
\citet{li-etal-2021-mrn} devise a novel mention-based reasoning module based on explicitly and collaboratively local and global reasoning.
\citet{DBLP:conf/ijcai/ZhangCXDTCHSC21} regard document-level RE as a semantic segmentation task and developed a document U-shaped network to capture both local context information and global interdependency among triples for document-level RE.

\subsection{Cross-Document Relation Extraction}
Earlier, some researchers probe into extracting entities, events, and relations from text in cross-document setting\citep{DBLP:conf/flairs/ZaraketM12, 10.1007/978-3-642-28604-9_25}. Recently, Cross-document Relation Extraction has been explored deeply by \citet{yao-etal-2021-codred}, who presents the first large-scale CodRE dataset, CodRED. 
To accomplish the task, \citet{yao-etal-2021-codred} propose two solutions, including a pipeline model and a joint model.
The pipeline method first extracts a relational graph for each document, and then reasons over these graphs to extract the target relation;
while the joint method directly aggregates different text path representations via a selective attention mechanism for the relation prediction.
We note that an effective CodRE system requires cross-document multi-hop reasoning through multiple potential bridging entities to narrow the semantic gap between documents.
However, the best-performing joint model in \citet{yao-etal-2021-codred} suffers from coarse-grained reasoning by merely synthesizing text paths in a shallow manner.
In this work, we consider modeling the global dependencies across multiple text paths (i.e., cross-path) based on bridging entities, which ensures more reliable reasoning for CodRE.

\section{Framework}

\textbf{Task Definition} 
Given a target entity pair $(e^h, e^t)$ and a bag of $N$ text paths $B=\left\{ p_i \right\}_{i=1}^N$, where each path $p_i$ consists of two documents $(d^h_i, d^t_i)$ mentioning the head entity $e^h$ and the tail entity $e^t$ separately, the task aims to infer the relation $r$ from $\mathcal{R}$ between the target entity pair, where $\mathcal{R}$ is a pre-defined relation type set. 
When multiple mentions of one entity (subject to entity ID) appear in two documents respectively, this entity is said to be shared by two documents. Note that the two documents in every path may share multiple entities $E^b_i = \{e^b_i\}^M_{i=1}$, in the following we call them bridge entities.

\noindent\textbf{System Overview}
As shown in Fig.\ref{fig:framework}, the model consists of four tiers.
First, an entity-based document-context filter receives text paths as inputs, where each of them is composed of two documents.
The filter removes less relevant sentences from the text paths and reorganizes the remaining sentences into more compact inputs for subsequent tiers.
Afterward, a BERT encoder yields the representations for tokens and entities.
Then the cross-path entity relation attention module builds a bag-level entity relation matrix for capturing the global dependencies between the entities and relations in the bag, and outputs the entity relation representations of all text paths.
Finally, we use a classifier to aggregate these representations and predict the relation between head and tail entities.

\subsection{Entity-based Document-context Filter}
\label{sec:entity_document_filter}

Since the average length of a document in CodRED is more than 4,900 tokens and BERT has a length limitation (512 tokens) for input, it is infeasible to handle all sentences in a text path simultaneously if the total length of all the input exceeds the limitation.
To solve this problem, we propose an entity-based document-context filter to select salient sentences in a document for each path. 

\begin{figure}[!t]
    \centering
    \includegraphics[width=0.95\columnwidth,height=0.28\textwidth]{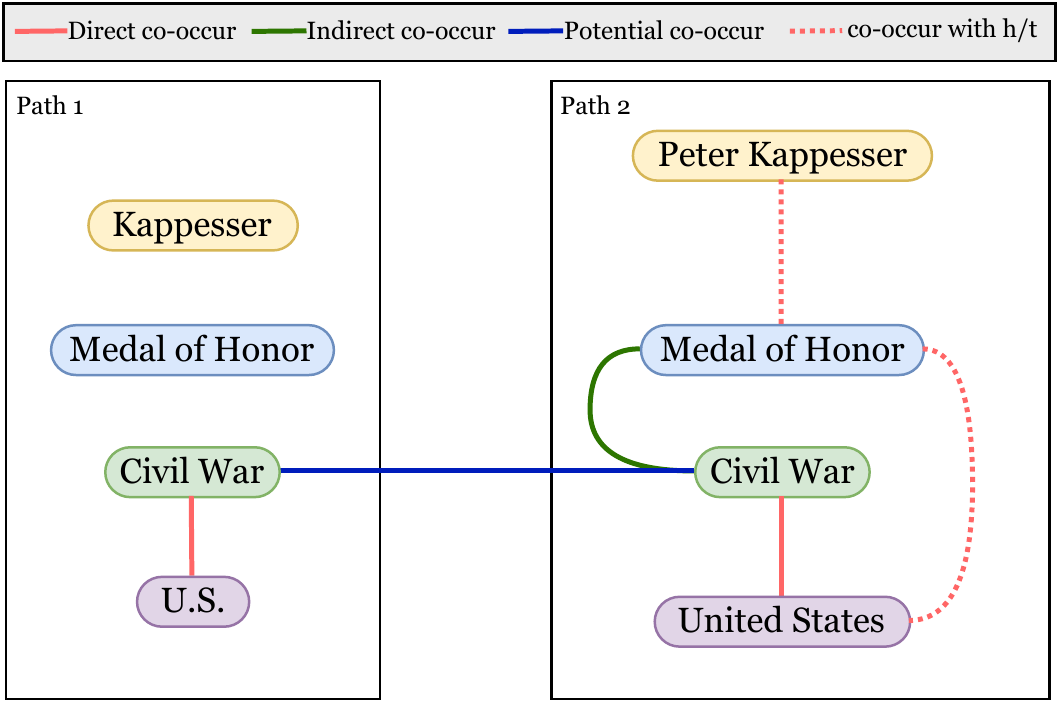}
    \caption{
    An example of the co-occurring graph for Path 1 and Path 2 in Fig.\ref{fig:example}. The score of ``\emph{Civil War}'' is obtained by aggregating the scores obtained from three conditions $\Gamma_1, \Gamma_2, \Gamma_3$ as shown in Equation (\ref{eq:score_e}).
    }
    \label{fig:co-occur}
    \vspace{-8pt}
\end{figure}

For each path $p$, we have a collection of entities $E^b$ shared by the two documents $(d^h, d^t)$ of this text path. These bridge entities can be utilized as a link in reasoning about the relation between head/tail entities. 
Moreover, the bridge entity collections can be regarded as a latent indicator to measure the distribution similarity between different text paths.
Thus, we first filter out a number of sentences based on their scores, which are computed by three heuristic conditions.
Then we use a semantic-based sentence filter to reorder the selected sentences to construct a coherent document whose length is less than 512.

\subsubsection{Entity-based Sentence Filtering}
\label{subsec:entity based sentence filtering}
The basic assumption of this module is that \textit{If a sentence includes entities that co-occur with a target entity, the sentence is informative for relation reasoning}. Thus our first filtering procedure is to select those informative sentences with prior distribution knowledge of bridge entities. To this end, we use three steps:

\textbf{Step 1:} We calculate the co-occurring score for each bridge entity. We design three heuristic conditions from strong to weak to describe the different levels of co-occurring situations: 
\begin{itemize}

    \item \textbf{Direct co-occur} $(\Gamma_1)$: Whether it co-occurs with the head/tail entity in the same sentence.
    \item \textbf{Indirect co-occur} $(\Gamma_2)$: Whether it co-occurs with another entity meets the first condition.
    \item \textbf{Potential co-occur} $(\Gamma_3)$: Whether it exists in other text paths.
\end{itemize}

Formally, for a bag of $N$ text paths, we score for each bridge entity $e^b$ in each text path $p_i$ by:
\begin{equation}
\setlength\abovedisplayskip{3pt}
\setlength\belowdisplayskip{3pt}
\label{eq:score_e}
    \text{score}(e^b) = \alpha s_1(e^b) + \beta s_2(e^b) + \gamma s_3(e^b)
\end{equation}
\begin{equation}
\setlength\abovedisplayskip{3pt}
\setlength\belowdisplayskip{3pt}
\begin{split}
s_1(e^b)= \left \{
\begin{array}{ll}
    1,                    & \text{if} \ \Gamma_1(e^b)\\
    0,                    & \text{otherwise}
\end{array}
\right.
\end{split}
\end{equation}
\begin{equation}
\label{eq:s2}
\setlength\abovedisplayskip{3pt}
\setlength\belowdisplayskip{3pt}
\begin{split}
s_2(e^b)\!=\! \left \{\!
\begin{array}{ll}
    |\left\{e^o|\Gamma_1(e^o)\land {\rm I}(e^o)=1\right\}|,\!\text{if} \ \Gamma_2(e^b)\\
    0,     \qquad \qquad \qquad \qquad \qquad   \text{otherwise}
\end{array}
\right.
\end{split}
\end{equation}
\begin{equation}
\label{eq:s3}
\setlength\abovedisplayskip{3pt}
\setlength\belowdisplayskip{3pt}
\begin{split}
s_3(e^b)= \left \{
\begin{array}{ll}
    |\left\{p_j| e^b \in {\rm E}_{j}^{b}\right\}|,  & \text{if} \ \Gamma_3(e^b)\\
    0,                      & \text{otherwise}
\end{array}
\right.
\end{split}
\end{equation}
where $\alpha, \beta, \gamma$ are hyper-parameters.
${\rm I}(e^o)=1$ while $e^o$ and $e^b$ co-occur in the same sentence, where $e^o \in E_i^b \setminus \left\{e^b\right\}$.
equation(\ref{eq:s2}) sums number of these $e^o$, equation(\ref{eq:s3}) sums number of these $p_j$.

\textbf{Step 2:} We compute the importance score $g^s$ of each sentence $s$ by summarizing all the scores of the bridge entities that it contains:
\begin{equation}
\setlength\abovedisplayskip{3pt}
\setlength\belowdisplayskip{3pt}
    g^s = \sum_{e^b \in {E}^{b}_s} \text{score}(e^b)
\end{equation}
where ${E}^{b}_s$ denotes the bridge entities mentioned in the sentence $s$.

\textbf{Step 3:} We rank the sentences by their importance scores from large to small and select the top $K$ sentences as the  candidate set $S=\left\{s_1,s_2,...,s_K \right\}$, where $K$ is a hyper-parameter. 
In our implementation, the candidate set size $K$ is set to 16 based on the experiments on the development set.
If there are several sentences with the same score, the priority is determined according to the distances from these sentences to the sentence with the highest score. 

\subsubsection{Semantic-based Sentence Filtering}
After the entity-based sentence filtering, we take the semantic relevance of sentences into account to further filter and reorder candidate sentences, with the assumption that \textit{if a sentence is semantically similar to the sentence including target entities, this sentence should be more informative for relation reasoning}. 
The goal of this step is to yield the most informative context $S^*$ from the candidate sentence set $S$, for reasoning the relation between target entities.

The procedure of semantic-based sentence filtering is summarized as Algorithm \ref{Alg: SentFilter}, which aims to construct the sequence $S^*$ from the candidate sentence set.
As seen, besides the candidate set $S$, head entity $h$ and tail entity $t$, the inputs of the algorithm also include a start set $S_{start}$ and an end set $S_{end}$ that consist of all the sentences containing the head and tail entity, respectively.
At the begging of the algorithm, we first randomly select a sentence from $S_{start}$ (line 1).
Then we search for the most relevant sentence to this sentence and append it to the output $S^*$.
We repeat such a process until the current selected sentence includes the tail entity (lines 3-12).
Finally, we obtain the sequence $S^*$ with $K^*$ sentences, where $K^* \leq K$.
Specifically, we use the cosine similarity calculated by SBERT-WK \cite{9140343} to measure the semantic relevance between two sentences.
If the length of the sequence $S^*$ is larger than 512, we will keep dropping the sentences with lower similarity scores until the length of the sequence meets the demand of BERT.

\begin{algorithm}[t]
	\caption{\textbf{Semantic-based Sentence Filtering}}
	\label{Alg: SentFilter}
	\begin{algorithmic}[1]
		\Require
		Candidate set $S=\left\{s_1,s_2,...,s_K \right\}$;
		head entity $h$; tail entity $t$; 
		Start set $S_{start}=\left\{s_i|h \in s_i \right\}$;
		End set $S_{end}=\left\{s_j|t \in s_j \right\}$;
		\Ensure 
		Sequential sentences $S^*$
		\State $S^*=\left[\quad \right]$; $cur={\rm Random}{(S_{start}, 1)}$
		\State $next=\phi$; $max=0$
		
		\State \textbf{while} $\left\{cur\right\} \cap S_{end} = \emptyset $ \textbf{do}
		\State \qquad $S = S-\left\{cur\right\}$
		\State \qquad \textbf{for} $s_i \in S$ \textbf{do}
		\State \qquad \qquad \textbf{if} {\rm Sim}($cur, s_i$) $> max$ 
        \State \qquad \qquad \qquad $max \gets$ {\rm Sim}($cur, s_i$)
        \State \qquad \qquad \qquad $next \gets$ $s_i$
        \State \qquad \qquad \textbf{else}
        \State \qquad \qquad \qquad continue
        \State \qquad $S^* = $ {\rm Append}($S^*, next$)
        \State \qquad $cur \gets next$
		\State \textbf{return} $S^*$
	\end{algorithmic}
\vspace{-3pt}
\end{algorithm}

\subsection{Encoder Module}
\label{sec:encoder}

After input construction, we have filtered sentence set $S^*$ from each text path,
we concatenate sentences in $S^*$ together to build the input of our model as $X=\left\{ w_{i} \right\}_{i=1}^L$.
Following \citet{yao-etal-2021-codred}, we apply unused tokens in the BERT vocabulary \citep{DBLP:conf/naacl/DevlinCLT19} to mark the start and end of every entity. Then we leverage BERT as the encoder to yield token representations:
\begin{equation}
\setlength\abovedisplayskip{3pt}
\setlength\belowdisplayskip{3pt}
    \left\{ \bm{h}_i \right\}_{i=1}^L={\rm BERT} (\left\{ w_{i} \right\}_{i=1}^L)
\end{equation}
Based on $\left\{ \bm{h}_{i} \right\}_{i=1}^L$, we can obtain the entity representations with the max-pooling operation:
\begin{equation}
\setlength\abovedisplayskip{3pt}
\setlength\belowdisplayskip{3pt}
    \bm{e}_j={\rm Max}\left\{\bm{h}_{i} \right\}_{j=start_j}^{end_j}
\end{equation}
where $start_j$ and $end_j$ are the start and end positions of the $j$-th mention.

\subsection{Cross-Path Entity Relation Attention}
\label{sec:cross_path_attention}
Since prior studies only treated each text path as an independent instance, the rich information across text paths was ignored. Therefore, we aim to mine this information. 
Inspired by \citet{DBLP:journals/corr/abs-2006-03719} and \citet{DBLP:conf/ijcai/ZhangCXDTCHSC21}, we introduce a cross-path entity relation attention module based on the Transformer \cite{NIPS2017_3f5ee243} to capture the inter-dependencies among the relations across paths. 

Concretely, we first collect all the entity mention representations in a bag and then generate relation representations for entity pairs:
\begin{equation}
\setlength\abovedisplayskip{3pt}
\setlength\belowdisplayskip{3pt}
    \bm{r}_{u,v}={\rm ReLU}({\bm{W}_r}({\rm ReLU}({\bm{W}_u}{\bm{e}_u}+{\bm{W}_v}{\bm{e}_v})))
\end{equation}
where $\bm{W}_r$, $\bm{W}_u$, $\bm{W}_v$ are learnable parameters.
Afterward, we extend the relation matrix proposed by \citet{DBLP:journals/corr/abs-2006-03719} at the bag level, as shown in Fig. \ref{fig:framework}(c).
In order to modeling the interaction among relations across paths,
we build a relation matrix $\bm{M} \in \mathbb{R}^{|E| \times |E| \times d}$, where $E=\bigcup_{i=1}^{N}E_i$ denotes all the entities in the entity set $E_i$ of text path $p_i$ and $E_i = \{e_i^h, e_i^t\} \cup E^b_i$.

To capture the intra- and inter-path  dependencies, we leverage a multi-layer Transformer \citep{NIPS2017_3f5ee243} to perform self-attention on the flattened relation matrix $\hat{\bm{M}} \in \mathbb{R}^{|E|^2 \times d}$:
\begin{equation}
\setlength\abovedisplayskip{3pt}
\setlength\belowdisplayskip{0pt}
    \hat{\bm{M}}^{(t+1)} = {\rm Transformer}(\hat{\bm{M}}^{(t)})
\end{equation}

Finally, we obtain the target relation representation ${\bm r}_{h_i,t_i}$ for each path $p_i$ from the last layer of the Transformer, as shown in Figure \ref{fig:framework}(c).

\subsection{Classifier}
Afterwards, we yield the relation representation ${\bm r}_{h_i,t_i}$ from each text path $p_i$ for each pair of target entities. 
Then we use the ${\bm r}_{h_i,t_i}$ as the classification feature and feed it into an MLP classifier for calculating the score of each relation:
\begin{equation}
\setlength\abovedisplayskip{3pt}
\setlength\belowdisplayskip{3pt}
    {\hat{y}}_i= {\rm MLP}({\bm r}_{h_i,t_i})
\end{equation}
To get the bag level prediction, we use the max-pooling operation on each relation label to yield the final score for each relation type $r$:
\begin{equation}
\setlength\abovedisplayskip{3pt}
\setlength\belowdisplayskip{3pt}
    {\hat{y}}^{(r)} = {\rm Max}\left\{  {\hat{y}}_i^{(r)} \right\}_{i=1}^N
\end{equation}

After obtaining the scores for all relations, we utilize a global threshold $\theta$, which will be stated in Section \ref{sec:loss fuction}, to filter out the categories lower than the threshold.

\subsection{Training Details}
\label{sec:loss fuction}
Since some bags have multiple relation labels, 
we adopt a multi-label global-threshold loss, which is a variant of the circle loss \cite{circleloss}, as our loss function.
To this end, we introduce an additional threshold to control which class should be output. 
We hope that the scores of the target classes are greater than the threshold and the scores of the non-target classes are less than the threshold. Formally, for each Bag $B$, we have:
\begin{equation}
\setlength\abovedisplayskip{3pt}
\setlength\belowdisplayskip{3pt}
\begin{split}
    \mathcal{L} = \ &log(e^\theta + \sum_{r\in \Omega_{neg}^B} e^{{\bm{\hat{y}}}^{(r)}}) \\
            &+log(e^{-\theta} + \sum_{r\in \Omega_{pos}^B} e^{{\bm{\hat{y}}}^{(r)}})
\end{split}
\end{equation}
where ${\bm{\hat{y}}}^{(r)}$ denotes the score for the relation $r$, 
$\theta$ denotes the threshold and is set to zero,
$\Omega_{pos}^B$ and $\Omega_{neg}^B$ are the positive and negative classes between the target entity pair.

\section{Experimental Settings}

\subsection{CodRED Dataset}
The CodRED dataset was constructed by \citet{yao-etal-2021-codred} from the English Wikipedia and Wikidata, which covers 276 relation types. 
The statistics of our data are shown in Table \ref{table:dataset statistics}, which is the same as that used in \citet{yao-etal-2021-codred}.

\begin{table}[t]
\centering
\small
\begin{tabular}{lccc}
\hline
    ~ & Train & Dev & Test\\
\hline
Bags (Pos) & 2,733 &  1,010 & 1,012\\
Bags (N/A)  & 16,668 & 4,558 & 4,523\\
Text paths & 129,548 & 40,740 & 40,524\\
Bridges & 613,566 & 195,766 & 197,888\\
Tokens/Doc & 4,938.6 & 5,031.6 & 5,129.2 \\
Path/Bag & 6.67 & 7.31 & 7.32\\
\hline
\end{tabular}
\caption{\label{table:dataset statistics}
Statistics of CodRED.
}
\end{table}

\begin{table*}[ht]
\centering
\small
\begin{tabular}{lcccc|cc}
\hline
\multirow{2}*{\textbf{Model}} & \multicolumn{4}{c}{\textbf{Dev}}& \multicolumn{2}{c}{\textbf{Test}}\\
\cline{2-5}\cline{6-7}
~ & F1 & AUC & P@500 & P@1000 & F1 & AUC\\
\hline
Pipeline \cite{yao-etal-2021-codred}  & 30.54 & 17.45 & 30.60 & 26.70 & 32.29 & 18.94\\
End-to-end \cite{yao-etal-2021-codred} & 51.26 & 47.94 & 62.80 & 51.00 & 51.02 & 47.46\\ \cdashline{1-7}
ECRIM (ours) & \textbf{61.12} & \textbf{60.91} & \textbf{78.89} & \textbf{60.17} & \textbf{62.48} & \textbf{60.67}\\
\hline
\end{tabular}
\caption{\label{main results}
Comparisons with the baselines on CodRED. The results of the baselines are extracted from the original paper. Our test results are obtained from the official website of CodRED on Codalab.
}
\vspace{-8pt}
\end{table*}

\subsection{Implementation Details and Evaluation Metrics}
We conduct our experiments using the closed setting of the benchmark dataset CodRED.\footnote{Available at \href{https://github.com/thunlp/CodRED}{https://github.com/thunlp/CodRED}.}
We use the cased BERT-base as the encoder. 
AdamW \citep{DBLP:conf/iclr/LoshchilovH19} is used to optimize the neural networks with a linear warm-up and decay learning rate schedule.
The learning rate is 3e-5, and the embedding and hidden dimension is 768. 
The $\alpha, \beta, \gamma$ in \ref{subsec:entity based sentence filtering} are 0.1, 0.01, 0.001 respectively. 
The Transformer encoder in \ref{sec:cross_path_attention} have 3 layers.
We tuned the hyper-parameters on the development set.
Other parameters in the network are all obtained by random initialization and updated during training.

Following \citet{yao-etal-2021-codred}, we adopt the F1/AUC/P@500/P@1000 (ignore N/A predictions) as the evaluation metrics for the experiments on the development set, and F1/AUC (ignore N/A predictions) for the experiments on the test set. Results are obtained from CodaLab.\footnote{ \href{https://codalab.lisn.upsaclay.fr/competitions/3770}{https://codalab.lisn.upsaclay.fr/competitions/3770}.}
For each target entity pair, the model yield a logit for each relation type. We rank $(h,t,r)$ according to the logit values from high to low, and select the top-N values to compute an average precision called P@N.
The F1/AUC/P@N (ignore N/A predictions) means that logits of $(h,t,n/a)$ will not be included in the calculation of F1/AUC/P@N.

\subsection{Baselines}
We compare our proposed model with two baselines provided by \citet{yao-etal-2021-codred}.

\noindent\textbf{Pipeline.}
\citet{yao-etal-2021-codred} build a pipeline model that decomposes cross-document RE into three phases: 
1) firstly, predicting the relations between the entities within a document to yield a relational graph containing head or tail entities;
2) secondly, for each entity $e$ shared by two relational graphs, predicting the relation $(h,e)$ and $(e,t)$ respectively, then concatenate the two relation representation and feed it into a fully connected layer to obtain relation distribution;
3) finally, aggregating the relation scores for all shared entity $e$ to obtain the final relation between the target entity pair.

\noindent\textbf{End-to-end.}
\citet{yao-etal-2021-codred} also design an end-to-end model to predict the relation. 
Specifically, they obtain representation for each text path $p_i$ by feeding tokens into BERT. 
Then they use selective attention mechanism to obtain an aggregated representation from all paths. 
Finally the aggregated representation is fed into a fully connected layer followed by a softmax layer to predict the relation between the entity pair.

\section{Results and Analyses}
\subsection{Main Results}

In this section, we report the main experimental results compared with the baseline models proposed by \citet{yao-etal-2021-codred}. 
As shown in table \ref{main results}, our model achieves superior performance in all metrics for both development set and test set. 
Specially, our method achieves 62.48\% F1 and 60.67\% AUC on the test set, and outperforms the best method End-to-end by 11.46\% and 13.21\% in terms of F1 and AUC.
The improvement in those scores verifies the excellent ability of our model due to our design for bridge entities and cross-path interaction. These two points will be further discussed in \ref{sec:effect analysis for bridges} and \ref{sec:effect analysis for paths}.


\subsection{Ablation Studies}

In this section, we conduct ablation experiments to verify the effectiveness of each component of our model.  We implement following model variants:

\begin{enumerate}[itemsep=0pt,parsep=0pt,label=(\arabic*)]
    \item ${\rm ECRIM}_{w/o\ IC}$, a variant that replaces the input construction module with the method used by \citet{yao-etal-2021-codred}, which evaluates the contribution of the input construction module.
    \item ${\rm ECRIM}_{w/o\ BR}$, a variant that discards bridge entities when constructing relation matrix, i.e. the relation matrix merely composed of the relations of target entities.
    \item ${\rm ECRIM}_{w/o\ CP}$, a variant that uses inner-path entity relation attention instead of cross-path entity relation attention, which is to validate the effectiveness of the utility of relation dependencies flow across text paths.
    \item  ${\rm ECRIM}_{w/o\ TH}$, a variant that replaces the threshold loss with the cross entropy loss.
\end{enumerate}

All these variants use the BERT-based as encoder. And the results are presented in Table \ref{ablation results}, from which we can observe that:

\begin{enumerate}[itemsep=0pt,parsep=0pt,label=(\arabic*)]
    \item The performance of ${\rm ECRIM}_{w/o\ IC}$ drops significantly, which confirms the importance of retaining significant information related to bridge entities.
    \item The result of ${\rm ECRIM}_{w/o\ BR}$ variant shows that when bridge entities were ablated, the performance of model declined substantially. This proves that relations with respect to bridge entities are very important.
    \item The performance of ${\rm ECRIM}_{w/o\ CP}$ model drops significantly as the cross-path entity relation attention module is discarded and replaced with Inner-Path Entity Relation Attention. This phenomenon indicates the effectiveness of enabling relations to interact with each other across text paths.
    \item The performance of ${\rm ECRIM}_{w/o\ TH}$ variant has decreased, demonstrating the effectiveness of the threshold loss we used.
\end{enumerate}

\begin{table}[t]
\centering
\small
\begin{tabular}{lcccc}
\hline
\textbf{Model} & F1 & AUC & P@500 & P@1000\\
\hline
ECRIM & 61.12 & 60.91 & 78.89 & 60.17\\ \cdashline{1-5}
\quad-IC & 59.14 & 59.16 & 75.78 & 58.46\\
\quad-BR & 56.99 & 57.85 & 73.21 & 56.80\\
\quad-CP & 58.63 & 59.27 & 74.36 & 57.62\\
\quad-TH & 60.22 & 59.80 & 75.42 & 58.74\\

\hline
\end{tabular}
\caption{\label{ablation results}
Ablation studies on the CodRED development set. 
IC: Input Construction; BR: Bridge Entity; CP: Cross-Path Entity Relation Attention; TH: Threshold.
}
\vspace{-8pt}
\end{table}

\subsection{Effect on the Number of Bridge Entities}
\label{sec:effect analysis for bridges}

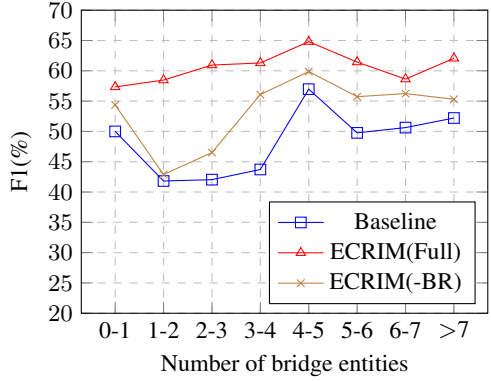
\begin{figure}[t]
    \centering
    \small
    \begin{tikzpicture}
        \begin{axis}[
            legend pos=south east,
            width=0.9\columnwidth,
            height=0.35\textwidth,
            xlabel={Number of bridge entities},
            ylabel={F1(\%)},
            ymin=20, ymax=70,
            xtick=data,
            symbolic x coords={0-1,1-2,2-3,3-4,4-5,5-6,6-7,$>$7},
            ytick={20,25,30,35,40,45,50,55,60,65,70},
            xmajorgrids=true,
            ymajorgrids=true,
            grid style=dashed,
            compat=newest,
        ]
        
        \addplot[
            color=blue,
            mark=square,
            ]
            coordinates {
                        (0-1,50.00)
                        (1-2,41.82)
                        (2-3,42.04)
                        (3-4,43.72)
                        (4-5,56.97)
                        (5-6,49.76)
                        (6-7,50.63)
                        ($>$7,52.21)
                        };
        \addplot[
            color=red,
            mark=triangle,
            ]
            coordinates {
                        (0-1,57.34)
                        (1-2,58.47)
                        (2-3,60.95)
                        (3-4,61.29)
                        (4-5,64.81)
                        (5-6, 61.44)
                        (6-7, 58.63)
                        ($>$7,62.06)
                        };
        \addplot[
            color=brown,
            mark=x]
            coordinates {
                        (0-1,54.36)
                        (1-2,42.89)
                        (2-3,46.51)
                        (3-4,56.08)
                        (4-5,59.88)
                        (5-6,55.73)
                        (6-7,56.24)
                        ($>$7,55.30)
                        };
        \legend{Baseline, ECRIM(Full), ECRIM(-BR)}
        \end{axis}
    \end{tikzpicture}
    \vspace*{0.5mm}
    \caption{
    The effect on F1s with regards to different numbers of bridge entities per path in bags.
    }
    \label{fig:bridges_ablation}
\end{figure}

\begin{figure}[t]
    \centering
    \small
    \begin{tikzpicture}
        \begin{axis}[
            legend pos=south east,
            width=0.9\columnwidth,
            height=0.35\textwidth,
            xlabel={Number of paths},
            ylabel={F1(\%)},
            ymin=30, ymax=70,
            xtick=data,
            symbolic x coords={1-4,5-8,9-12,13-16,$>$16,$>$20},
            ytick={30,35,40,45,50,55,60,65,70},
            xmajorgrids=true,
            ymajorgrids=true,
            grid style=dashed,
            compat=newest,
        ]
        
        \addplot[
            color=blue,
            mark=square,
            ]
            coordinates {
                        (1-4,40.65)
                        (5-8,33.52)
                        (9-12,62.79)
                        (13-16, 63.08)
                        ($>$16, 65.51)
                        };
        \addplot[
            color=red,
            mark=triangle,
            ]
            coordinates {
                        (1-4,48.10)
                        (5-8,49.87)
                        (9-12,67.45)
                        (13-16, 69.19)
                        ($>$16, 68.18)
                        };
        \addplot[
            color=brown,
            mark=x]
            coordinates {
                        (1-4,38.27)
                        (5-8,36.04)
                        (9-12,66.13)
                        (13-16, 67.81)
                        ($>$16, 65.39)
                        };
        \legend{Baseline, ECRIM(Full), ECRIM(-CP)}
        \end{axis}
    \end{tikzpicture}
    \vspace*{0.5mm}
    \caption{
    The effect on F1s with regards to different numbers of paths in bags.
    }
    \vspace{-8pt}
    \label{fig:paths_ablation}
\end{figure}
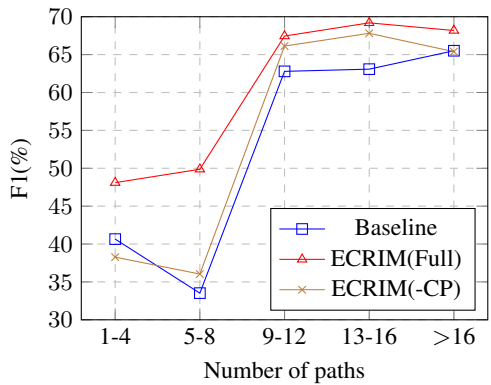

\begin{figure}[t]
    \centering
    \includegraphics[width=1.0\columnwidth,height=0.42\textwidth]{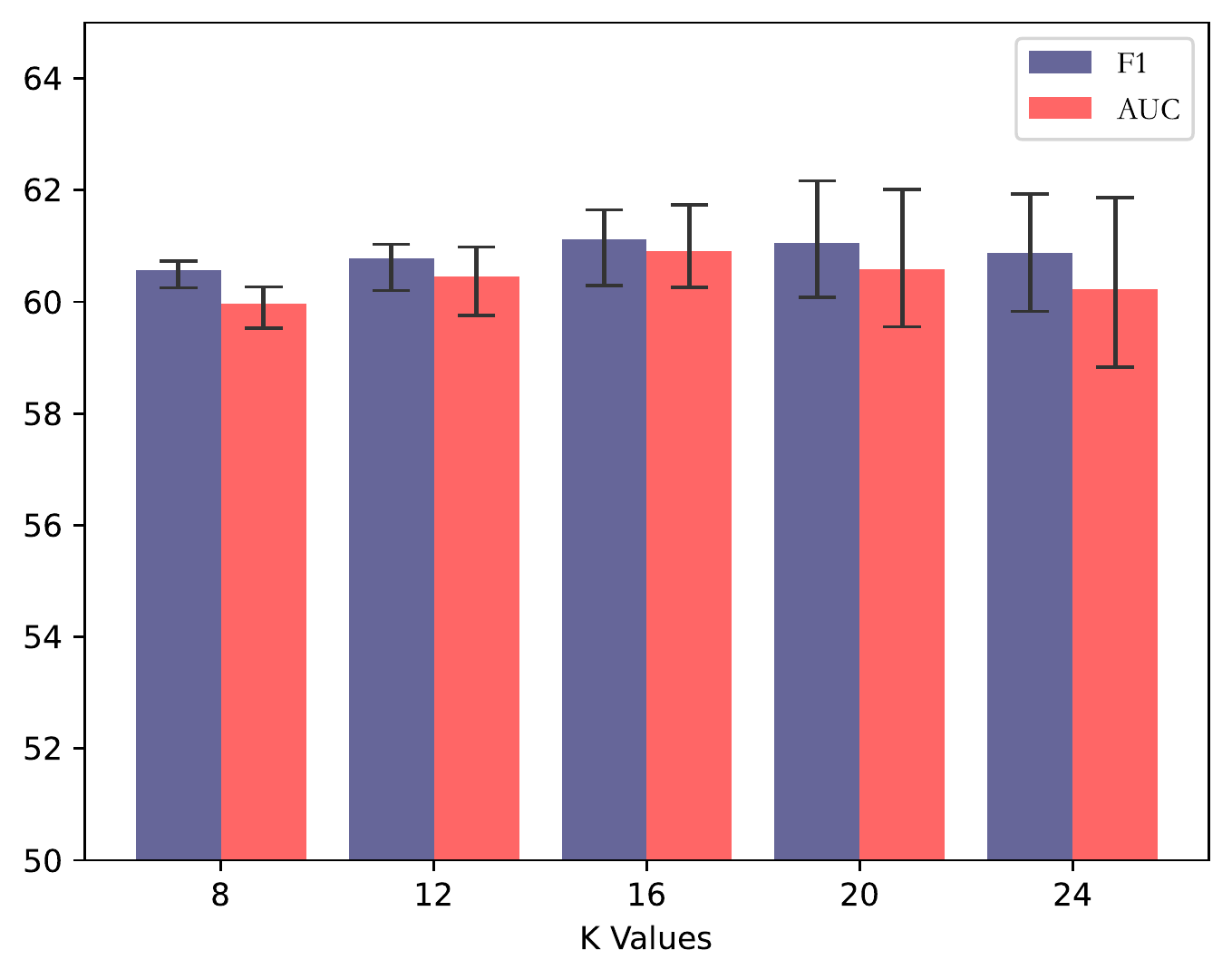}
    \caption{The effect on F1s and AUCs with offset under the different selections of $K$ in Section \ref{sec:entity_document_filter}.
    }
    \vspace{-8pt}
    \label{fig:K_offset}
\end{figure}

To investigate the effect of bridge entities for cross-document relation extraction, we divided the origin dev set of CodRED into several subsets by the average number of bridges per path in a bag. We report the model performance on these subsets as shown in Fig.\ref{fig:bridges_ablation}. We can observe that as the number of bridge entities increases, the performance of the model increases first because the bridge entities bring more information shared by the two documents in the path. This evidently proves the necessity to utilize the bridge entity information for cross-document RE.

As the number of bridge entities in the path continues to increase, the performance of the model decreases slightly. This is due to the increase in noise caused by a large number of bridge entities. The complex context makes the reasoning process of the model difficult. However, our model is better than baseline in resisting this noise, as our model can distinguish noise factors in finer granularity.

\subsection{Effect Analysis for the Number of Paths}
\label{sec:effect analysis for paths}

To investigate the impact of path numbers within a bag for CodRE, we divided the origin dev set of CodRED into several subsets by path numbers of each bag. We report the model performance on these subsets as shown in Fig.\ref{fig:paths_ablation}. We can observe that all models perform better with a larger number of paths than with a small number of paths, as the number of positive paths in a bag also increases. 
Our model ECRIM (Full) achieves a great improvement compared to baseline when the number of paths is small, which shows that when facing difficult situations with fewer paths, our model makes full use of cross-path information for reasoning.

\subsection{Which is the appropriate value for hyperparameter $K$?}
\label{sec:Analysis for Hyperparameter K}

In this section, we experiment on the development set to heuristically search for the appropriate value of hyperparameter $K$. The influence of $K$ value is mainly in two aspects: constraint for the uncertainty brought by algorithm \ref{Alg: SentFilter} and constraint the computational time cost for the execution of algorithm \ref{Alg: SentFilter}. Figure.\ref{fig:K_offset}. shows that with the increase of $K$ value, the fluctuation degree of model performance is greater. 
On the other hand, table \ref{table:K complexity} shows the significant impact of $K$ value on the algorithm execution time. 
Considering the effects of both aspects, we set the value of $K$ to 16 to ensure that the fluctuation is small when the time cost is acceptable.

\begin{table}[t]
\small
    \centering
        \begin{tabular}{lcc}
        \hline
         & Speed($\rm{bags} \cdot \rm{min}^{-1}$)\\
        \hline
        $K=8$ & 37.89 \\
        $K=12$ & 15.43 \\
        $K=16$ & 8.28 \\
        $K=20$ & 4.72 \\
        $K=24$ & 2.53 \\
        \hline
        \end{tabular}
    \caption{\label{table:K complexity}
    Comparison of algorithm \ref{Alg: SentFilter} execution speed under different K value settings.
    }
    
\end{table}

\begin{table}[t]
\small
    \centering
        \begin{tabular}{lccc}
        \hline
        MBE & MUA & TMU & ST\\
        \hline
        2 & 1940 & 10859 & 5.03\\
        3 & 2580 & 11499 & 5.06\\
        4 & 3630 & 12549 & 5.01\\
        5 & 5256 & 14175 & 5.03\\
        6 & 7822 & 16741 & 4.93\\
        7 & 11418 & 20337 & 4.87\\
        8 & 15334 & 24253 & 4.67\\
        >8 & - & Out of Memory & -\\
        \hline
        \end{tabular}
    \caption{\label{table:computation cost}
    GPU memory usage and running speed for different number of bridge entities. MBE denotes Max number of Bridge Entities per path, MUA denotes Memory Usage (MiB) of Attention matrix, TMU denotes Total Memory Usage (MiB) of Model and ST represents the Speed ($\rm{bags} \cdot \rm{min}^{-1}$) on the Train set. 
    }
    \vspace{-8pt}
\end{table}

\subsection{Case Study}
\begin{figure}[t]
    \centering
    \includegraphics[width=1.0\columnwidth,height=0.42\textwidth]{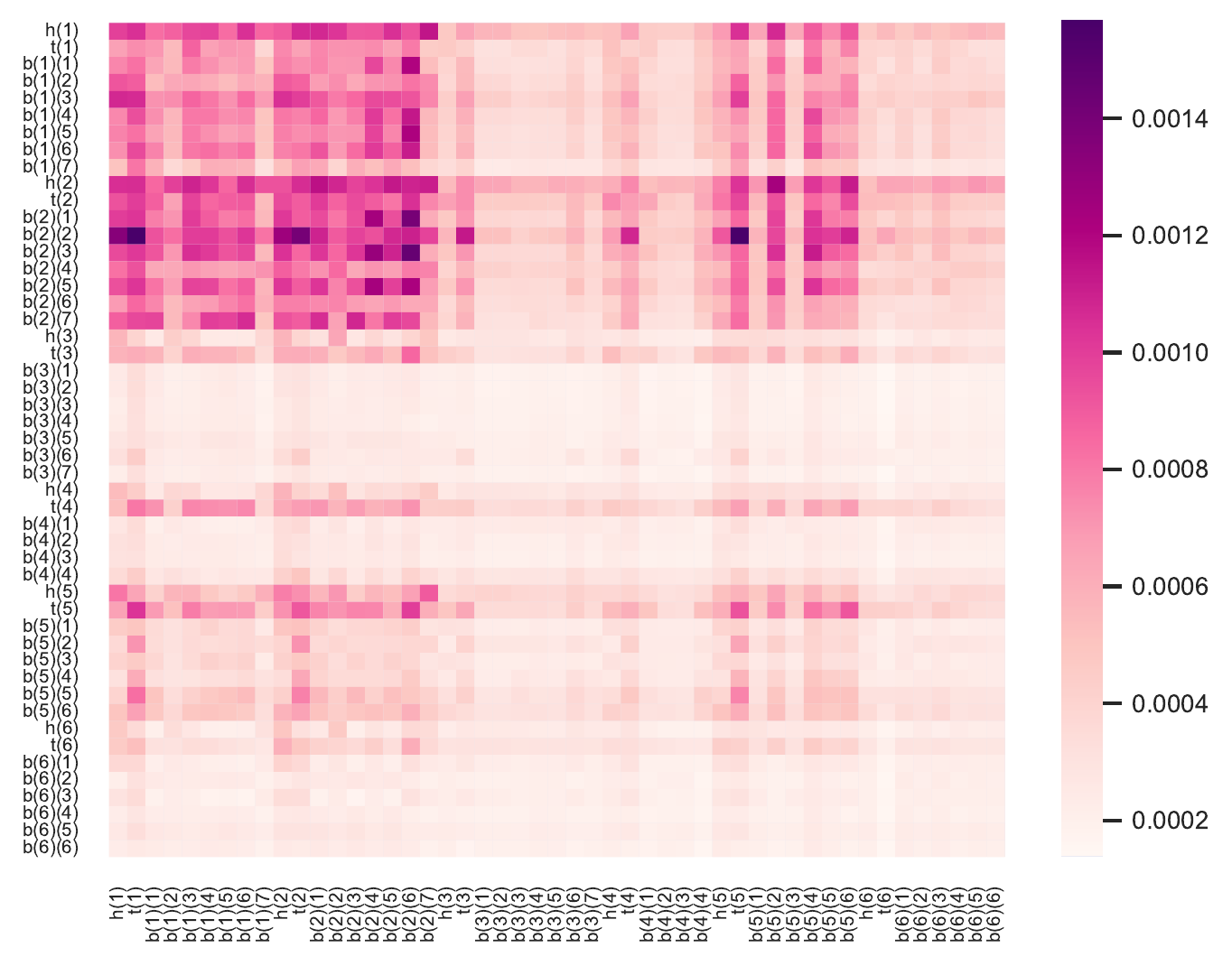}
    \caption{A case study to show the normalized attention scores of a target relation unit $(h_1,t_1)$ for all the relations in the bag.
    }
    \vspace{-8pt}
    \label{fig:heatmap}
\end{figure}

To further illustrate the effectiveness of cross-path dependency between relations learned by our model, we present a case study which can be seen in Fig.\ref{fig:heatmap}, an attention score matrix heatmap of relation $(h_1, t_1)$. 
For example, the unit in row 3 and column 2 (whose coordinate is $(t(1),b(1)(1))$) represents the contribution of the relation between the tail entity of $p_1$ and the first bridge entity of $p1$ to $r_{h_1,t_1}$, where $t(i)$ represents the tail entity of the $i$-th path, and $b(i)(j)$ denotes the $j$-th bridge entity of the $i$-th path. 
Obviously, the most prominent areas on the heatmap are the four blocks in the upper left corner, which denote the inner relation of $p_1$ and $p_2$ and the cross path relation between $p_1,p_2$. As the ground-truth label of these paths is $[P126, P126, n/a, n/a, n/a, n/a]$ ($P126$ and $n/a$ refers to \emph{relation ID} and \emph{no relation}), it is proved that the model successfully learns the cross-path dependency which contributes to the prediction.

\section{Conclusions}
\label{sec:bibtex}
In this paper, we devise an entity-based document-context filter to extract important snippets related to the target information for cross-document RE. For relation prediction, we propose a model that considers the global dependencies across multiple text paths and performs a fine-grained reasoning process simultaneously. Empirical results show that our method drastically improves the performance of cross-document relation extraction.
Our work can be a valuable reference for this research.

\section*{Limitations}
Because the model is built at the bag level, the computational complexity of cross-path entity relation attention will grow with the increasing numbers of text paths and bridge entities, resulting in an increase in GPU memory demand and a decrease in inference speed. As table \ref{table:computation cost} demonstrated, the usage of GPU memory increases rapidly with the increase of bridge entity number, as the shape of the Attention Matrix is ${\rm{|E|}^4} \times d$. Meanwhile, the computing efficiency decrease slightly.
If there are too many potential paths, we have to discard some of them to maintain the feasibility of our model.

In addition, the entity-based document-context filter that we use to construct the input is unsupervised and not learnable. 
How to build a learnable model to extract more informative sentences from long documents is a future work that has much room for exploration.
Another potential line is to explicitly model the discourse structure of the relevant documents, over which the reasoning of the RE or cross-document RE will be easier.


\section*{Acknowledgements}
We thank all the reviewers for their insightful comments. This work is supported by the National Natural Science Foundation of China (No. 62176187), the National Key Research and Development Program of China (No. 2017YFC1200500), the Research Foundation of Ministry of Education of China (No. 18JZD015), the Youth Fund for Humanities and Social Science Research of Ministry of Education of China (No. 22YJCZH064), the General Project of Natural Science Foundation of Hubei Province (No.2021CFB385).

\bibliography{anthology,custom}
\bibliographystyle{acl_natbib}


\end{CJK}
\end{document}